\documentclass[conference]{IEEEtran}
\IEEEoverridecommandlockouts
\usepackage{cite}
\usepackage{amsmath,amssymb,amsfonts}
\usepackage{algorithmic}
\usepackage{graphicx}
\usepackage{textcomp}
\usepackage[table,dvipsnames]{xcolor}
\usepackage{graphics}
\usepackage{adjustbox}
\usepackage{graphicx}
\usepackage{threeparttable}
\usepackage{multirow}

\def\BibTeX{{\rm B\kern-.05em{\sc i\kern-.025em b}\kern-.08em
    T\kern-.1667em\lower.7ex\hbox{E}\kern-.125emX}}

\usepackage{enumitem}    

\newif\ifoutline
\outlinetrue

\colorlet{mc}{LimeGreen!50!White!50!}
\newcommand{\hlc}{\cellcolor{mc}}

\renewcommand{\IEEEbibitemsep}{0pt plus 0.5pt}
\makeatletter
\IEEEtriggercmd{\reset@font\normalfont\fontsize{7.9pt}{8.60pt}\selectfont}
\makeatother
\IEEEtriggeratref{1}

\makeatletter
\newcommand*\titleheader[1]{\gdef\@titleheader{#1}}
\AtBeginDocument{%
  \let\st@red@title\@title
  \def\@title{%
    \bgroup\normalfont\normalsize\centering\@titleheader\par\egroup
    \vskip1ex\st@red@title}
}
\makeatother

\titleheader{\vspace{-30pt}To appear at the 25th Design, Automation and Test in Europe Conference (DATE'22), Mar 14-23 2022, Antwerp, Belgium.}
\title{Cross-Layer Approximation For Printed \\Machine Learning Circuits}

\begin{document}
\bstctlcite{IEEEexample:BSTcontrol} 
\setlength{\abovedisplayskip}{2ex}
\setlength{\belowdisplayskip}{2ex}

\author{\IEEEauthorblockN{
Giorgos Armeniakos\IEEEauthorrefmark{1}\IEEEauthorrefmark{2},
Georgios Zervakis\IEEEauthorrefmark{2},
Dimitrios Soudris\IEEEauthorrefmark{1},
Mehdi B. Tahoori\IEEEauthorrefmark{2},
and J{\"o}rg Henkel\IEEEauthorrefmark{2}
}
\IEEEauthorblockA{\IEEEauthorrefmark{1}National Technical University of Athens, Greece,
\IEEEauthorrefmark{2}Karlsruhe Institute of Technology, Germany}
\IEEEauthorblockA{
\IEEEauthorrefmark{1}\{armeniakos, dsoudris\}@microlab.ntua.gr,
\IEEEauthorrefmark{2}\{georgios.armeniakos, georgios.zervakis, mehdi.tahoori, henkel\}@kit.edu
}
}

\maketitle

\begin{abstract}
Printed electronics (PE) feature low non-recurring engineering costs and low per unit-area fabrication costs, enabling thus extremely low-cost and on-demand hardware.
Such low-cost fabrication allows for high customization that would be infeasible in silicon, and bespoke architectures prevail to improve the efficiency of emerging PE machine learning (ML) applications.
However, even with bespoke architectures, the large feature sizes in PE constraint the complexity of the ML models that can be implemented.
In this work, we bring together, for the first time, approximate computing and PE design targeting to enable complex ML models, such as Multi-Layer Perceptrons (MLPs) and Support Vector Machines (SVMs),  in PE.
To this end, we propose and implement a cross-layer approximation, tailored for bespoke ML architectures.
At the algorithmic level we apply a hardware-driven coefficient approximation of the ML model and at the circuit level we apply a netlist pruning through a full search exploration.
In our extensive experimental evaluation we consider 14 MLPs and SVMs and evaluate more than 4300 approximate and exact designs.
Our results demonstrate that our cross approximation delivers Pareto optimal designs that, compared to the state-of-the-art exact designs, feature 47\% and 44\% average area and power reduction, respectively, and less than 1\% accuracy loss.
\end{abstract}


\begin{table*}[t!]
\setlength\tabcolsep{3pt}
\caption{Evaluation of Bespoke Printed ML Circuits.}
\label{tab:baselines}
\footnotesize
\centering
\renewcommand{\arraystretch}{1.1}
\begin{threeparttable}
\begin{tabular}{l|ccccc|ccccc|ccccc|ccccc}
\hline
 & \multicolumn{5}{c|}{\textbf{MLP-C}} & \multicolumn{5}{c|}{\textbf{MLP-R}} & \multicolumn{5}{c|}{\textbf{SVM-C}} & \multicolumn{5}{c}{\textbf{SVM-R}} \\ \hline
 & Acc\tnote{1}  & T\tnote{2} & \#C\tnote{3}  & \begin{tabular}[c]{@{}c@{}}Area \\ ($cm^{2}$)\end{tabular} & \begin{tabular}[c]{@{}c@{}}Power\\ ($mW$)\end{tabular} & Acc\tnote{1}  & T\tnote{2} & \#C\tnote{3} & \begin{tabular}[c]{@{}c@{}}Area \\ ($cm^{2}$)\end{tabular} & \begin{tabular}[c]{@{}c@{}}Power\\ ($mW$)\end{tabular} & Acc\tnote{1}  & T\tnote{2} & \#C\tnote{3}  & \begin{tabular}[c]{@{}c@{}}Area \\ ($cm^{2}$)\end{tabular} & \begin{tabular}[c]{@{}c@{}}Power\\ ($mW$)\end{tabular} & Acc\tnote{1}  & T\tnote{2} & \#C\tnote{3} & \begin{tabular}[c]{@{}c@{}}Area \\ ($cm^{2}$)\end{tabular} & \begin{tabular}[c]{@{}c@{}}Power\\ ($mW$)\end{tabular} \\ \hline
\textbf{Cardio}    & 0.88 & (21,3,3)  & 72  & 33.4 & 97.3 & 0.83 & (21,3,1)  & 66 & 21.6 & 65.9 & 0.90 & 3  & 63  & 15.1 & 46.8 & 0.84 & 1  & 21 & 6.8  & 22.9 \\
\textbf{Pendigits} & 0.94 & (16,5,10)  & 130 & 67.0 & 213.0    & 0.37 & (16,5,1)  & 85 & -\tnote{4}     & -    & 0.98 & 45 & 160 & 123.8 & 364.4    & 0.23 & 1  & 16 & -\tnote{4}     & -    \\
\textbf{RedWine}   & 0.56 & (11,2,6)  & 34  & 17.6 & 53.3 & 0.56 & (11,2,1)  & 24 & 7.1  & 24.0 & 0.57 & 15 & 66  & 23.5 & 72.9 & 0.56 & 1  & 11 & 4.0  & 15.1 \\
\textbf{WhiteWine} & 0.54 & (11,4,7)  & 72  & 31.2 & 98.4 & 0.53 & (11,4,1)  & 48 & 13.1 & 40.7 & 0.53 & 21 & 77  & 28.3 & 87.4 & 0.53 & 1  & 11 & 4.2  & 15.5 \\ \hline
\end{tabular}
\begin{tablenotes}\footnotesize
\item[] $^1$ Accuracy using $8$-bit coefficients and $4$-bit inputs.
$^2$ Model's topology (for SVMs: the number of classifiers). $^3$ Number of coefficients of the model. $^4$ These models achieve low accuracy and are not evaluated.
\vspace{-2ex}
\end{tablenotes}
\end{threeparttable}
\end{table*}

\section{Introduction}\label{sec:intro}

Printed electronics emerge as a promising solution for application domains such as smart packaging, disposables (e.g., packaged foods, beverages), fast moving consumer goods (FMCG), in-situ monitoring, low-end healthcare products (e.g., as smart bandages) etc~\cite{Mubarik:MICRO:2020:printedml}.
Such domains have requirements for ultra-low cost and conformality that silicon-based systems cannot satisfy. For example, increased manufacturing, packaging, and assembly costs of silicon prevent sub-cent costs while silicon systems cannot meet either stretchability, porosity, and flexibility requirements~\cite{Bleier:ISCA:2020:printedmicro}.

Although, printed circuits promise to satisfy these demands through their ultra-low cost additive manufacturing that enables conformal hardware on-demand, the large feature sizes in printed electronics mean that the associated hardware overheads will be prohibitive for complex circuits.
The latter include implementations of machine learning (ML) classification algorithms that are required by a large number of printed applications in aforementioned domains~\cite{Mubarik:MICRO:2020:printedml}.
Hence, works on printed ML classifiers are limited.

In~\cite{Mubarik:MICRO:2020:printedml}, the authors exploit i) the hardware efficiency of bespoke architectures~\cite{Kumar2017Bespoke} and ii) the fact that the non-recurring engineering and low fabrication costs of printed electronics enable on-demand printing of ML circuits customized to a specific model and implement bespoke printed classifiers, paving the way for ML classification even on printed electronics.
Such degree of customization is mainly infeasible in silicon systems.
However, due to area and power constraints,~\cite{Mubarik:MICRO:2020:printedml} implemented only simple ML algorithms such as Decision Trees and Support Vector Machine Regression (SVM-Rs).

In this paper, we  investigate the possibility of implementing more complex printed ML circuits and specifically Multi-Layer Perceptrons classifiers (MLP-Cs), Multi-layer Perceptron regressors (MLP-Rs), SVM classifiers (SVM-C), as well as SVM-Rs in printed electronics. 
To this end, to further improve the area efficiency compared to the state-of-the-art bespoke ML implementations~\cite{Mubarik:MICRO:2020:printedml}, we adopt, for the first time, Approximate Computing (AC) principles in printed electronics.
Approximate computing exploits the intrinsic error resilience of a large number of application domains, such as ML, and trades computational accuracy for gains in other metrics, such as area and power~\cite{Shafique:DAC:2016:cross}.
Designing approximate arithmetic circuits has gained a vast research interest~\cite{Honglan:JPROC:2020:axsurv,Waris:TCASII:2020}.
Moreover, to mitigate the increased complexity of approximate design, several works focus on approximate design automation~\cite{ApproxSynthSurvey2020,Barbar2021ApplicationDriven,Mrazek:DAC:2019:autoax,Zervakis2019MultiLevel}.
Notably, significant research interest is shown in gate-level pruning approaches~\cite{GatePrun2017,Scarabottolo:DAC:2019:prune}
due to its inherent efficiency in reducing a circuit's area complexity.
Gate-level pruning is a circuit agnostic technique that operates on an already optimized netlist and further reduces its area by removing selected gates.
Though, netlist pruning state of the art mainly targets arithmetic circuits only.
In addition, research activities on approximate bespoke circuits are very limited.

In this work, we leverage cross-layer approximation, that is proven to outperform single layer (either logic or algorithmic) techniques~\cite{Shafique:DAC:2016:cross,Zervakis2019MultiLevel}, and we propose and implement an automated cross-approximation framework customized for approximate bespoke ML circuits.
At the algorithmic level, our framework applies coefficient approximation in which the coefficients of a given trained model are replaced with more (bespoke) hardware-friendly values.
At the logic level, we implement a netlist pruning approach that is customized for printed bespoke architectures and through a full search design space exploration it extracts Pareto-optimal solutions.
Using our framework, we elucidate the impact of approximate computing on designing complex printed ML circuits.
We demonstrate that, compared to the state-of-the-art exact printed ML circuits, our framework decreases the area (power) by $47$\% ($44$\%) on average. 
In many cases, these gains are sufficient to enable complex printed ML circuits.
It is noteworthy that our framework requires $12$min on average.
The latter is critical for printed electronics due to their on-demand and point-of-use--even at low to moderate volumes--fabrication process.

\noindent
\textbf{Our novel contributions within this work are as follows}:
\begin{enumerate}[topsep=0pt,leftmargin=*]
    \item This is the first work that evaluates the impact of approximate computing on printed electronics specifically when targeting printed ML circuits. 
    \item We propose, for the first time, an automated, cross-layer approximation framework for bespoke ML circuits.
    \item Using our framework, we demonstrate that, in many cases, approximate computing can be used to integrate complex ML models in printed circuits \footnote{Our implementations are available at https://github.com/garmeniakos/Ax-Printed-ML-Classifiers.git}
    
\end{enumerate}

\section{Background}\label{sec:background}

Printed electronics denotes a set of printing methods which can realize ultra low-cost \cite{subramanian2008printed}, large area \cite{chen201430} and flexible \cite{mohammed2017all} computing systems in combination with functional materials  to realize transistors and passive components on various substrates. There are different processes for the fabrication of printed circuits, such as screen printing,  jet-printing  or roll-to-roll processes \cite{de2010fully}. All these printing techniques refer to an additive manufacturing process, where functional materials are directly deposited on the substrate, which simplifies the production chain compared to subtractive silicon-based processes substantially \cite{chang2014fully}. This leads to savings in per unit-area costs and enables flexible hardware form factors.

Printed electronics do not compete with silicon-based electronics in terms
of integration density, area and performance. Typical frequencies achieved by
printed circuits are from a few Hz to a few
kHz~\cite{cadilha2017digital}. Similarly, the feature
size tends to be several
microns~\cite{lei2019low}. However, due to its form-factor, conformity and most importantly, significantly reduced fabrication costs -- even for low quantities -- it can target application domains, unreachable by conventional silicon-based VLSI. 

Despite these attractive features, there are several limitations of printed electronics compared to traditional silicon technologies. Due to large feature sizes, the integration density of printed circuits is orders of magnitude lower than silicon VLSI. Additionally, due to large intrinsic transistor gate capacitances, the performance of printed circuits is orders of magnitude lower compared to nanometer technologies. 

\section{Cross-Layer Approximation For Printed ML Bespoke Architectures}\label{sec:approximate}
In this section we present our automated cross-layer approximation framework for generating approximate printed ML circuits.
Briefly, our framework receives as input a trained model (e.g., dumped from scikit-learn) and performs a hardware-driven coefficient approximation (algorithmic level).
Then, it automatically generates the bespoke RTL description of the approximated model and synthesizes the circuit using the Electronic Design Automation (EDA) tool and printed Process Design Kit (PDK).
Finally, on the synthesized netlist, it applies logic approximation through gate-level pruning.

\subsection{Bespoke ML Architectures}\label{subsec:bespoke}

As aforementioned, printed electronics enable bespoke circuits customized per ML model~\cite{Mubarik:MICRO:2020:printedml}.
In such customized implementations, the coefficients are hardwired in the circuit leading to high area efficiency.
Hence, as a baseline, we consider an area-optimized fully-parallel bespoke circuit for each ML model examined, following the design methodology of~\cite{Mubarik:MICRO:2020:printedml}.
Specifically, we examine MLP-C, MLP-R, SVM-C, and SVM-R models trained on the Cardiotocography, Pendigits, RedWine, and WhiteWine datasets of the UCI ML repository~\cite{Dua:2019:uci}.
Training is performed using scikit-learn and the randomized parameter optimization (RandomizedSearchCV) with $5$-fold cross validation.
Inputs are normalized to $[0,1]$ and training/testing uses a random $70$\%/$30$\% split.
For the MLPs one hiden layer with up to five neurons and Relu activiation function are used.
SVMs use a linear kernel and SVM-Cs are implemented with 1-vs-1 classification.
We set the topology of each MLP so that each MLP features the least number of hidden nodes and all MLPs achieve close to maximum accuracy.
To generate each bespoke circuit, the coefficients and intercepts are hardwired in the Register-Transfer Level (RTL) description and fixed-point arithmetic is used.
The precision for the coefficients and inputs is set to $8$ and $4$ bits, respectively, since these values delivered close to floating-point accuracy for all the models.
The RTL descriptions are synthesized using Synopsys Design Compiler and mapped to the open source Electrolyte Gated Transistor (EGT) library~\cite{Bleier:ISCA:2020:printedmicro}, which is an inkjet-printed technology.
EGT is low-voltage and allows battery powered printed circuits.
All the circuits are synthesized at a relaxed clock (i.e., $250$ms for Pendigits MLP-C and $200$ms for the rest circuits) targeting to further improve the area efficiency.
These delay values are in compliance with typical performance of printed electronics~\cite{cadilha2017digital}.
Circuit simulations are performed with Questasim and the test dataset to obtain the circuit's switching activity.
Then, power analysis is performed using Synopsys PrimeTime and the previously obtained switching activity.
The aforementioned EDA tool-flow is used throughout our paper.

Table~\ref{tab:baselines} reports the characteristics (e.g., area, power) for our baseline bespoke implementations of all the considered models.
As shown, the area of almost all the circuits ($>20$cm$^2$) is prohibitive for most printed applications.
Similarly, their power consumption is so high ($>30$mW) that they cannot be powered by a single existing printed battery.
Only SVM-Rs and the RedWine MLP-R exhibit acceptable area and power.

\subsection{Hardware-Driven Coefficient Approximation}\label{subsec:weightax}

\begin{figure}[t!]
\centering
\includegraphics[]{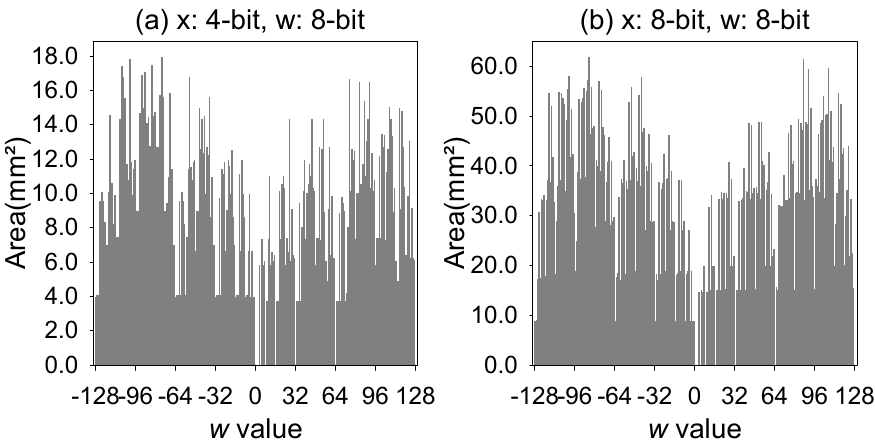}
\caption{The area of the bespoke multiplier w.r.t. the coefficient value $w$. Two architectures are considered: a) 4-bit inputs and 8-bit coefficients and b) 8-bit inputs and 8-bit coefficients.
For reference the area of the conventional $4\times 8$ and $8\times 8$ multipliers is 83.61$mm^{2}$ and 207.43$mm^{2}$, respectively.
}
\vspace{-2ex}
\label{fig:mult8area}
\end{figure}

A weighted sum (as for example example in the case of MLPs and SVMs) is expressed as:
\begin{equation}
    S=\sum_{1\leq i \leq N}{x_i\cdot w_i},
\end{equation}
where $w_i$ are the predefined coefficients (or weights) obtained after training, $x_i$ are the inputs, and $N$ is the number of coefficients.
In the case of bespoke ML architectures, these coefficients are hardwired within the circuit~\cite{Mubarik:MICRO:2020:printedml}.
As a result, the area (and power) of each bespoke multiplier $\mathrm{BM}_w$ required to compute the product $x\cdot w$, $\forall x$, is determined by the value of the coefficient $w$ and the width of the input $x$.
For example, Fig.~\ref{fig:mult8area} presents the area of $\mathrm{BM}_w$, $\forall w \in [-128,127]$ (i.e., 8-bit coefficients), for 4-bit and 8-bit input values.
For completeness, in the caption of Fig.~\ref{fig:mult8area} we also report the area of the conventional $4\times 8$ and $8 \times 8$ multipliers.
In both cases, the bespoke multipliers $\mathrm{BM}_w$ feature significantly lower area than the conventional multiplier for all the $w$ values.
Moreover, it is evident that the area of $\mathrm{BM}_w$ highly depends on $w$ and the input bitwidth.
However, similar trend is observed in Fig~\ref{fig:mult8area}a and~\ref{fig:mult8area}b, i.e., neighbouring $w$ values may feature significantly different area.
Importantly, in many cases the area may be nullified, e.g., when $w$ is a power of two.
This motivates us to investigate and propose a hardware-driven coefficient approximation, tailored for bespoke architectures, that replaces a coefficient value $w$ with a neighbouring value $\tilde{w}$ so that \texttt{AREA}($\mathrm{BM}_{\tilde{w}}$) $<$ \texttt{AREA}($\mathrm{BM}_w$).

\begin{figure}[t!]
\centering
\includegraphics[width=0.45\columnwidth]{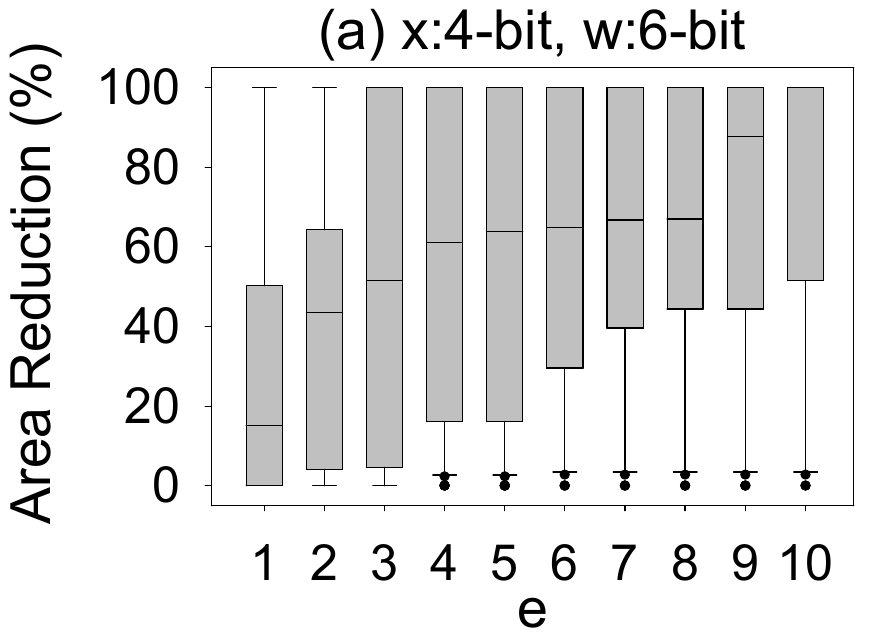}
\includegraphics[width=0.45\columnwidth]{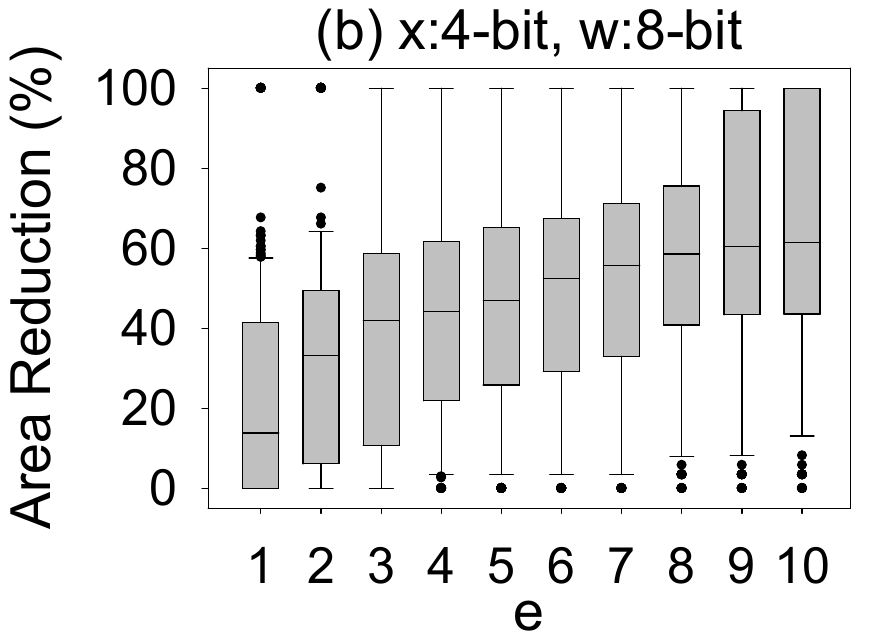}\\
\vspace{1ex}
\includegraphics[width=0.45\columnwidth]{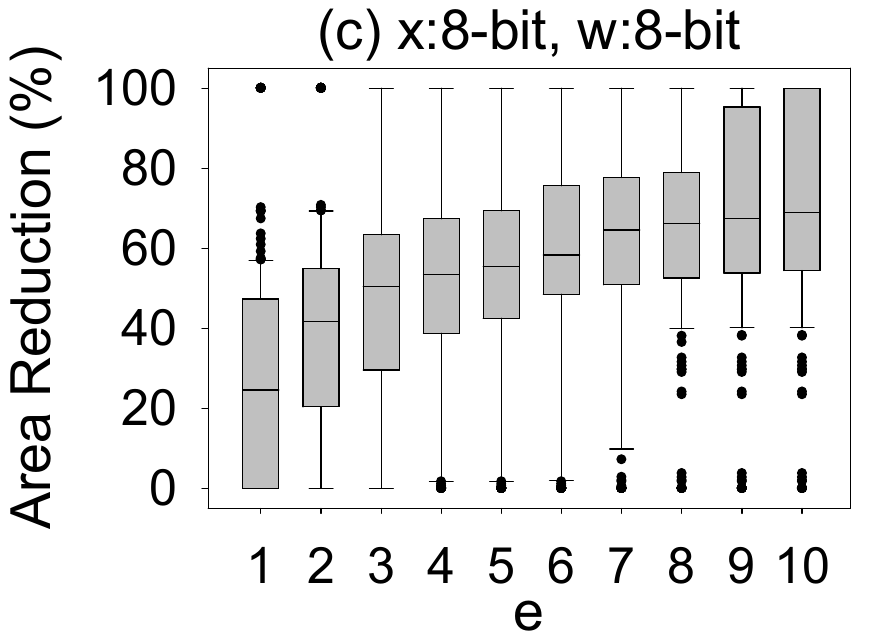}
\includegraphics[width=0.45\columnwidth]{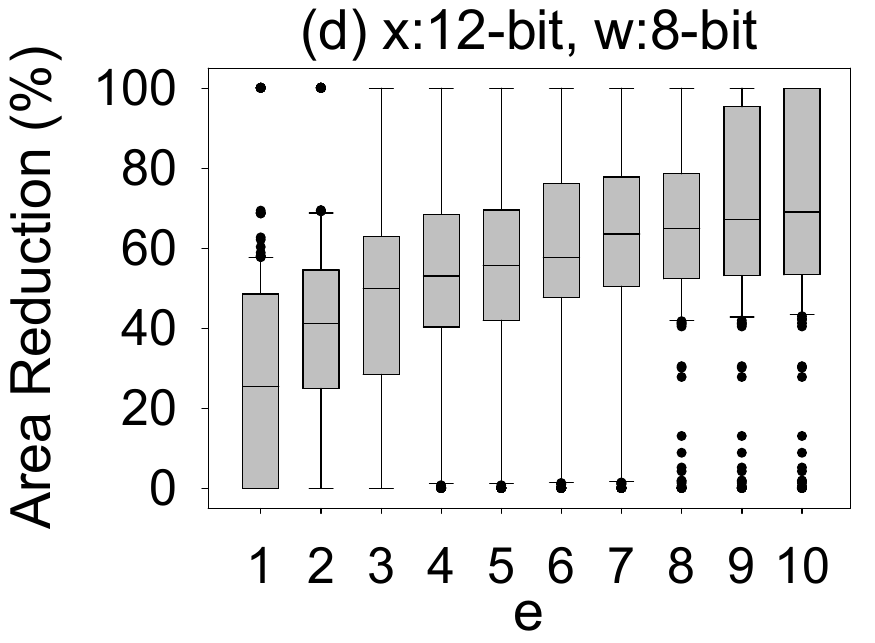}
\vspace{-2ex}
\caption{The area reduction delivered by our coefficient approximation when $(w-\tilde{w})\leq e$. Several bespoke multipliers are considered (a-d).}\vspace{-13pt}
\label{fig:multareasav}
\end{figure}

Fig.~\ref{fig:multareasav} presents the area reduction that is achieved by our coefficient approximation with respect to several bespoke multipliers sizes (a-d).
To generate each boxplot in Fig.~\ref{fig:multareasav}, for all $w$, we select $\tilde{w}$ so that $\tilde{w}$ features the lowest \texttt{AREA}($\mathrm{BM}_{\tilde{w}}$) and $\tilde{w} \in [w-e,w+e]$, where $e$ is a given threshold (x-axis).
Clipping is applied at the borders.
As shown in Fig.~\ref{fig:multareasav}, the obtained $\mathrm{BM}_{\tilde{w}}$ feature significantly lower area than the $\mathrm{BM}_w$.
Our coefficient approximation delivers a median area reduction of more than $19\%$ when $e=1$ while this value increases to $53\%$ when $e=4$.
Nevertheless, in most cases, for $e\geq4$ the area reduction becomes less significant.
For example, in Fig.~\ref{fig:multareasav}b, the median area reduction is $44\%$ for $e=4$ and increases to only $61\%$ for $e=10$.
In many cases, in Fig.~\ref{fig:multareasav}, the area reduction goes up to $100$\% or it is $0$\%.
The former is explained by the fact that $w$ was replaced by $\tilde{w}$ that was a power of two and thus the area reduction is $100$\% since $\tilde{w}$ features zero area.
On the other hand, in the cases that $w$ features the lowest area in the segment $[w-e,w+e]$, $w$ is not replaced and the area reduction is zero.

When replacing $w$ by $\tilde{w}$, the multiplication error is equal to $x\cdot(w-\tilde{w})$.
Thus, the error $\epsilon_S$ of the weighted sum is:
\begin{equation}\label{eq:wsumerror}
    \epsilon_S =\sum_{1\leq i \leq N}{x_i\cdot (w_i-\tilde{w}_i) }
\end{equation}
Considering positive inputs (see Section \ref{subsec:bespoke}), by systematically selecting $\tilde{w}_i$ to balance the positive and negative errors (i.e., $w_i-\tilde{w}_i$), we can minimize~\eqref{eq:wsumerror}.

Given an MLP or SVM, we implement our proposed hardware-driven coefficient approximation as follows:
\begin{enumerate}[leftmargin=*]
\item Given the coefficients $w_i$ and the bitwidth of the inputs, we evaluate the area of all the bespoke multipliers (\texttt{AREA}($\mathrm{BM}_{\tilde{w}}$)), $\forall i$ and $\forall \tilde{w} \in [w_i-e, w_i+e]$.
This step uses Synopsys Design Compiler and the EGT PDK~\cite{Bleier:ISCA:2020:printedmicro} for circuit synthesis and area analysis.~\label{item:synthbm}
\item For all the coefficients $w_i$, create a set $R_i=\{\tilde{w}_i^-,\tilde{w}_i^+\}$ s.t. $\tilde{w}_i^- \in [w, w+e]$ and  $\tilde{w}_i^-$ features the smallest area in that segment, i.e., \texttt{AREA}($\mathrm{BM}_{\tilde{w}_i^-}$) = min(\texttt{AREA}($\mathrm{BM}_{\tilde{w}_i}$)), $\forall \tilde{w} \in [w, w+e]$.
Similarly, we select $\tilde{w}_i^+ \in [w-e, w]$. By definition replacing $w$ with $\tilde{w}_i^-$ generates a negative error while replacing $w$ with $\tilde{w}_i^+$ generates a positive error.~\label{item:minarea}
\item We perform a brute-force search to select the configuration $\{\tilde{w}_i: \tilde{w}_i \in R_i, \forall i \}$ so that $\sum_{\forall i}{(w_i-\tilde{w}_i)}$ is minimized. In case of a tie, we select the one that minimizes $\sum_{\forall i}{\texttt{AREA}(\mathrm{BM}_{\tilde{w}_i})}$.~\label{item:miner}
\end{enumerate}
Note that steps~\ref{item:synthbm}-\ref{item:miner} are executed for each weighted sum, i.e., neuron in MLPs and 1-vs-1 classifier in SVMs.
In addition, we set $e$=$4$ in our analysis since for $e>4$ the area gains quite saturate (see Fig.~\ref{fig:multareasav}).
At the worst case, step~\ref{item:synthbm} required less than $6$s using $12$ threads (i.e., limit of available licenses).
In step~\ref{item:miner} we implement an exhaustive search to extract the final configuration.
Unlike conventional silicon VLSI, in printed electronics the examined ML models are rather small in size (in terms of number of parameters).
Hence, each weighted sum (neuron or classifier) features only a limited number of coefficients, i.e., the size of the design space is well constrained.
It is noteworthy, that in the worst case, step~\ref{item:miner} required only 5s using $80$ threads.
The aforementioned execution times refer to a dual-CPU Intel Xeon Gold 6138 server. 
In our optimization (steps~\ref{item:synthbm}-\ref{item:miner}) the sum $\sum_{\forall i}{\texttt{AREA}(\mathrm{BM}_{\tilde{w}_i})}$ is used as a proxy of the area of the weighted sum.
In other words, by minimizing the area (through our coefficient approximation) of the required bespoke multipliers, we aim in minimizing the area of the weighted sum.
We evaluate our area proxy against $1000$ randomly generated weighted sum circuits (i.e., random coefficients and input sizes).
The Pearson correlation coefficient between the area of the weighted sum obtained by Design Compiler and the area estimation using $\sum_{\forall i}{\texttt{AREA}(\mathrm{BM}_{\tilde{w}_i})}$ is $0.91$, i.e., perfect linear correlation.
Hence, our proxy precisely captures the area trend and minimizing $\sum_{\forall i}{\texttt{AREA}(\mathrm{BM}_{\tilde{w}_i})}$ in our optimization, will result in a weighted sum circuit with minimal area.
Finally, since our technique replaces the coefficient values with approximate more hardware-friendly ones, it does not require any specific/custom hardware implementation (e.g., as usually done in logic approximation).
Hence, it can be seamlessly integrated in any design framework and exploits all the optimization and IPs (e.g., multipliers) of synthesis tools.

\subsection{Netlist Pruning}\label{subsec:prune}

To further increase the area efficiency, in addition to our coefficient approximation, we apply netlist pruning.
Netlist pruning is based on the observation that the output of several gates in a netlist remains constant (`0' or `1') for the majority of the execution time.
Hence, removing such a gate from the netlist and replacing its output with a constant value, results in low error rate.
Netlist pruning has been widely studied to enable circuit-agnostic approximation~\cite{GatePrun2017,Scarabottolo:DAC:2019:prune}. 
In this section we provide a brief description of how we implemented and tailored netlist pruning for bespoke printed ML architectures.
First we define two pruning parameters for a gate: $\tau$ is the maximum percentage of time that the gate's output is `0' or `1' and $\phi$ the most significant output bit (starting from 0) that the gate is connected to (through any path).
Using $\tau$ and $\phi$ we can constraint the error frequency and the error magnitude, respectively. 
For example, assume that gate U1 is `1' the $\tau$=$90$\% of the time and that $\phi$=$3$.
Replacing U1 by `1' will result in an error rate of $10$\% and the maximum error will be less than $2^4$.
Netlist pruning is mainly implemented using heuristics~\cite{GatePrun2017} and thus optimality cannot be guaranteed.
In our work, leveraging that i) bespoke architectures feature significantly fewer area/gates than conventional architectures and ii) that ML models for printed electronics are rather limited in size, we use $\tau$ and $\phi$ to constraint the pruning design space and we implement an exhaustive search to obtain Pareto-optimal solutions.
Aiming for high area-efficiency, we prune all gates that feature $\tau$ and $\phi$ less or equal to given constraints $\tau_c$ and $\phi_c$.
Since all the pruned gates feature $\phi \leq \phi_c$, the maximum output error is less than $2^{\phi_c+1}$ irrespective of the number of pruned gates.
Overall, our coarse-grained approach ensures maximum area reduction while satisfying a maximum error threshold and enables fast design space exploration since only a gate's $\tau$ and $\phi$ need to be calculated.

Leveraging $\phi$ we filter all the gates that feature high $\tau$ and prune those that satisfy a given worst-case error.
In the case of regressors (MLP-R and SVM-R) this works well and many gates are pruned for low $\phi_c$ values.
However, classifiers require special consideration.
MLP-C and SVM-C use an argmax function at the end to translate the numerical predictions (e.g., values of output neurons in MLP-C) to a class.
As a result, the paths passing from all the gates are eventually congested in a few output bits, limiting the pruning granularity (possible $\phi_c$). 
Moreover, argmax breaks the correlation between the introduced numerical error in predictions and the final output.
For example, argmax([$0.9$, $0.1$])=argmax([$0.4$, $0.1$])=$0$.
For this reason, for the classifiers, we calculate $\phi$ for each gate with respect to the inputs of the argmax function.
For example, assume an MLP-C with $k$ output neurons $O_1$,..., $O_k$.
We define the value $\phi$ of a gate as $\max\limits_{\forall i}\phi(O_i)$, where $\phi(O_i)$ is the most significant output bit of the neuron $O_i$ that the gate is connected to.
If such a path doesn't exist, we set $\phi(O_i)=-1$.

Given a netlist (either exact or coefficient approximated), our netlist pruning operates as follows:
\begin{enumerate}[leftmargin=*]
\item Run RTL simulation of the synthesized netlist using the training dataset and Questasim to obtain the switching activity interchange format (SAIF) file.\label{item:sim}
\item Parse SAIF to calculate $\tau$ and the respective constant value (`0' or `1' ) for each gate.
For example, if the output of a gate is the $85$\% of the time `1' and the $15$\% it is `0', then $\tau$=$85$\% and the constant value is `1'. \label{item:tau}
\item Extract all the gates with $\tau \leq \tau_c$ and calculate their $\phi$. $\phi$ is easily calculated with the synthesis tool by reporting paths from a gate to the outputs.
\label{item:tauphi}
\item Prune all gates with $\tau \leq \tau_c$ and $\phi \leq \phi_c$, i.e., replace their output with the constant value extracted in step~\ref{item:tau}.\label{item:prune}
\item Synthesize the pruned netlist and evaluate its area and power as well as its accuracy on the test dataset.\label{item:synth}
\end{enumerate}
The pruned netlist is synthesized to exploit all optimizations of the synthesis tool, e.g., constant propagation.
Steps~\ref{item:sim} and~\ref{item:tau} are executed only once.
Step~\ref{item:tauphi} is executed $\forall \tau_c \in [80\%, 99\%]$.
Then, for each $\tau_c$, steps~\ref{item:prune}-\ref{item:synth} are executed $\forall \phi_c \in \Phi_\tau$.
$\Phi_\tau$ is equal to the set of the unique $\phi$ values obtained in step~\ref{item:tauphi}.
$\Phi_\tau$ enables us to explore only the relevant $\phi_c$ values, accelerating our full search exploration.
For example, if all the gates that feature $\tau\geq99$\% affect only the zero and first output bits, then $\phi_c>1$ is meaningless since it will return the same solution as $\phi_c=1$.
Unlike the pruning state of the art that examines only very simple circuits~\cite{GatePrun2017,Scarabottolo:DAC:2019:prune}, our implementation is evaluated on complex ML circuits.
Moreover, as explained above for the classifiers, conventional pruning~\cite{GatePrun2017,Scarabottolo:DAC:2019:prune} cannot be used.

\section{Results and Analysis}\label{sec:experimental}

\begin{figure}[t!]
\centering
\includegraphics[]{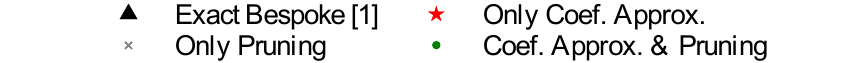}\\
\vspace{1ex}

\includegraphics[]{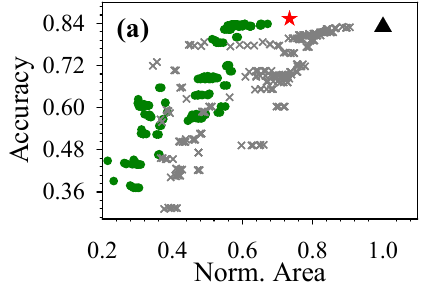}
\includegraphics[]{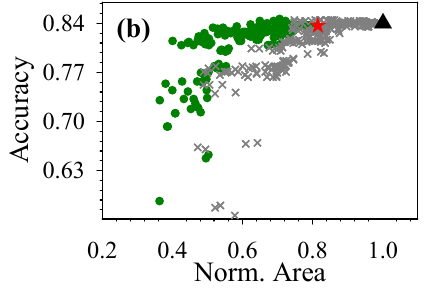}\\
\includegraphics[]{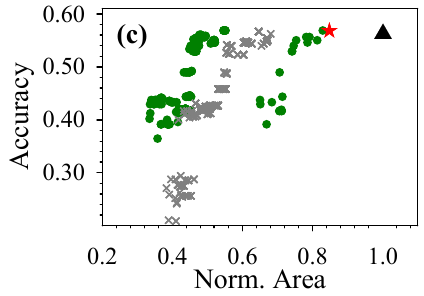}
\includegraphics[]{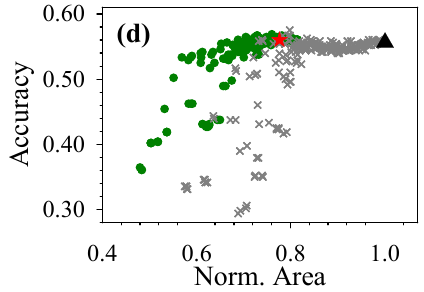}\\
\includegraphics[]{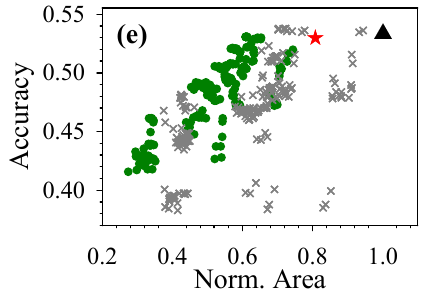}
\includegraphics[]{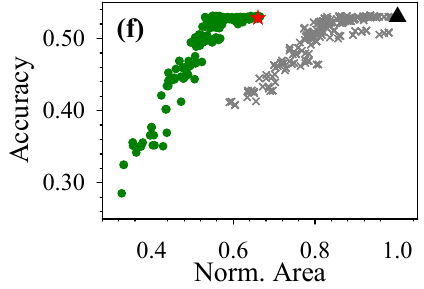}\\
\includegraphics[]{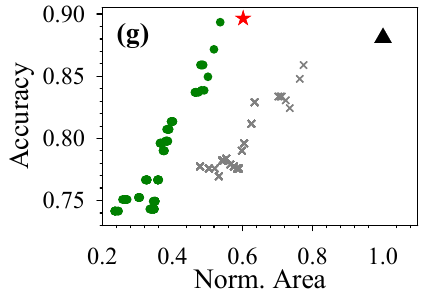}
\includegraphics[]{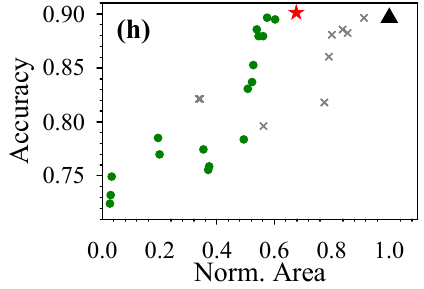}\\
\includegraphics[]{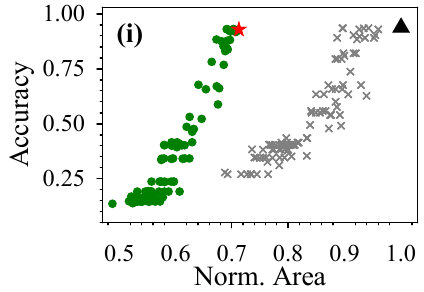}
\includegraphics[]{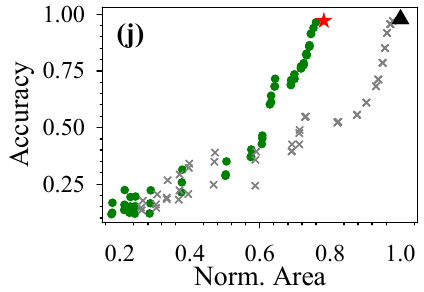}\\
\includegraphics[]{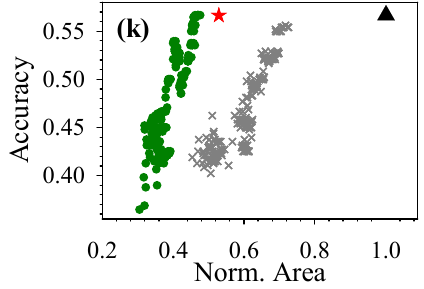}
\includegraphics[]{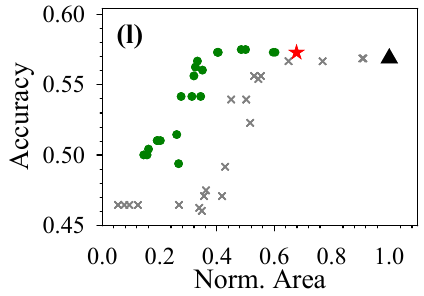}\\
\includegraphics[]{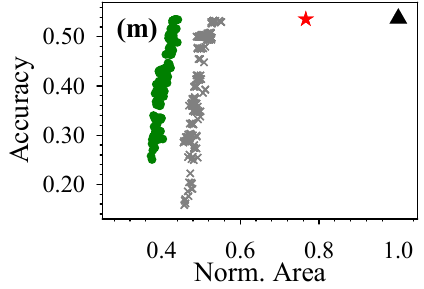}
\includegraphics[]{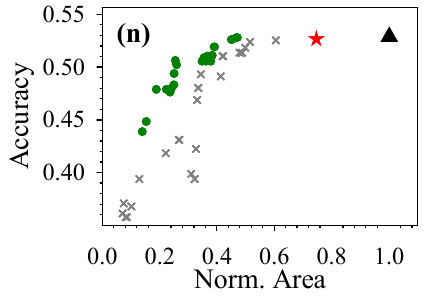}\\
\caption{
Accuracy vs normalized area Pareto space. ML models examined:
Cardio MLP-R (a),  SVM-R(b), MLP-C (g), and  SVM-C (h).
Pendigits MLP-C (i) and SVM-C (j).
RedWine MLP-R (c),  SVM-R(d), MLP-C (k), and  SVM-C (l).
WhiteWine MLP-R (e),  SVM-R(f), MLP-C (m), and  SVM-C (n).
}
\label{fig:dse}
\vspace{-4ex}
\end{figure}

In this section we evaluate the efficiency of our framework in reducing the area complexity and we investigate the impact of approximate computing on printed ML circuits.
Fig.~\ref{fig:dse} presents the Accuracy-Area Pareto space for all the printed ML circuits examined (see Table~\ref{tab:baselines}).
Area is reported as a normalized value w.r.t. the area of the respective baseline bespoke circuit. Identical results are obtained for the power consumption.
Due to space limitations, we present the area reduction since it is our primary optimization goal in printed circuits.
In Fig.~\ref{fig:dse}, the black triangles are the baseline bespoke designs.
The red star is the design that applies only our proposed coefficient approximation.
The green dots are the designs that are generated by our framework (i.e., Coefficient Approximation \& Pruning).
In addition, Fig.~\ref{fig:dse} also presents the designs that apply only pruning (gray `x').
To generate these designs, we apply our pruning method directly on the baseline circuit. 
In other words, the green dots are pruned derivatives of the red start while the gray `x' are pruned derivatives of the black triangle.
At this point note that our pruning method performs a full search exploration.
However, the number of the explored designs is circuit dependent.
As aforementioned, our pruning framework uses $80$\% $\leq \tau_c \leq$ $99$\% for all circuits but the explored $\phi_c$ values are circuit and $\tau_c$ dependent.
In total, we evaluated more than $4300$ designs.

Overall, it is observed that approximate computing can effectively be employed to decrease the area complexity of the printed ML circuits since all the approximate designs feature lower area than the exact one (i.e., black triangle).
For example, for most circuits, more than $57$\% area reduction can be achieved for less than $5$\% accuracy loss.
As shown in Fig.~\ref{fig:dse}, our coefficient approximation achieves $28$\% average area reduction for almost identical accuracy with the exact baseline and in most cases outperforms significantly the standalone gate-level pruning approximation.
In addition, the designs generated by our cross-layer approximation framework (green dots) constitute mainly the optimal solutions since they mainly form the Pareto front in all the subfigures of Fig.~\ref{fig:dse}.

\begin{table}[t]
\caption{Area and Power evaluation for less than $1$\% accuracy loss. Highlighted designs can be power by a Molex $30$mW battery.}
\label{tab:threshold}
\centering
\setlength\tabcolsep{2pt}
\footnotesize
\begin{threeparttable}
\begin{tabular}{l|cccc@{\hskip 4pt}|cccc@{\hskip 4pt}|cccc}
\hline
\multirow{2}{*}{\textbf{ML Circuit}}& \multicolumn{4}{c|}{\textbf{\begin{tabular}[c]{@{}c@{}}Coeff. Approx.\\ \& Pruning\end{tabular}}} & \multicolumn{4}{c|}{\textbf{\begin{tabular}[c]{@{}c@{}}Only\\Coeff. Approx.\end{tabular}}} & \multicolumn{4}{c}{\textbf{\begin{tabular}[c]{@{}c@{}}Only\\Pruning\end{tabular}}} \\ \cline{2-13}
\rule{0pt}{8pt}
& \begin{tabular}[c]{@{}c@{}}\textbf{A\tnote{\tiny 1}}\\  \end{tabular} & \begin{tabular}[c]{@{}c@{}}\textbf{P\tnote{\tiny 2}}\\  \end{tabular} & \textbf{\begin{tabular}[c]{@{}c@{}}AG\tnote{\tiny 3}\\ \end{tabular}} & \textbf{\begin{tabular}[c]{@{}c@{}}PG\tnote{\tiny 3}\\ \end{tabular}} & \begin{tabular}[c]{@{}c@{}}\textbf{A\tnote{\tiny 1}}\\  \end{tabular} & \begin{tabular}[c]{@{}c@{}}\textbf{P\tnote{\tiny 2}}\\  \end{tabular} & \textbf{\begin{tabular}[c]{@{}c@{}}AG\tnote{\tiny 3}\\ \end{tabular}} & \textbf{\begin{tabular}[c]{@{}c@{}}PG\tnote{\tiny 3}\\ \end{tabular}} & \begin{tabular}[c]{@{}c@{}}\textbf{A\tnote{\tiny 1}}\\  \end{tabular} & \begin{tabular}[c]{@{}c@{}}\textbf{P\tnote{\tiny 2}}\\  \end{tabular} & \textbf{\begin{tabular}[c]{@{}c@{}}AG\tnote{\tiny 3}\\ \end{tabular}} & \textbf{\begin{tabular}[c]{@{}c@{}}PG\tnote{\tiny 3}\\ \end{tabular}} \\ \hline

\textbf{Card MLP-R}  & 12 & 37 & 45 & 44 & 16 & 49 & 27 & 26 & 18 & 56 & 16 & 15 \\
\textbf{Card SVM-R}  & \hlc3.5 & \hlc13 & \hlc49 & \hlc42 & \hlc5.5 & \hlc19 & \hlc19 & \hlc15 & \hlc5.0 & \hlc18 & \hlc26 & \hlc22 \\
\textbf{RW MLP-R}  & \hlc3.3 &\hlc 12 & \hlc53 & \hlc49 & \hlc6.0 & \hlc21 & \hlc15 & \hlc14 & \hlc4.6 & \hlc17 & \hlc35 & \hlc30 \\
\textbf{RW SVM-R}  & \hlc2.6 & \hlc10 & \hlc35 & \hlc33 & \hlc3.1 & \hlc12 & \hlc22 & \hlc22 & \hlc2.9 & \hlc11 & \hlc27 & \hlc25 \\
\textbf{WW MLP-R}  & \hlc8.0 & \hlc27 & \hlc39 & \hlc35 & 11 & 34 & 20 & 17 & 9.2 & 29 & 30 & 28 \\
\textbf{WW SVM-R}  & \hlc2.2 & \hlc8.5 & \hlc47 & \hlc45 & \hlc2.8 & \hlc11 & \hlc34 & \hlc32 & \hlc3.4 & \hlc13 & \hlc19 & \hlc19 \\
\textbf{Card MLP-C}  & 17 & 54 & 48 & 44 & 20 & 62 & 40 & 36 & 33 & 97 & 0 & 0 \\
\textbf{Card SVM-C}  & \hlc8.7 & \hlc29 & \hlc43 & \hlc38 & 10 & 33 & 33 & 29 & 14 & 43 & 8.7 & 8.3 \\
\textbf{Pend MLP-C}  & 46 & 153 & 31 & 28 & 48 & 143 & 29 & 33 & 60 & 194 & 10 & 9.0 \\
\textbf{Pend SVM-C}  & 97 & 287 & 22 & 21 & 97 & 287 & 22 & 21 & 121 & 357 & 2.2 & 1.8 \\
\textbf{RW MLP-C}  & \hlc8.0 & \hlc27 & \hlc55 & \hlc50 & 9.3 & 30 & 47 & 43 & 18 & 53 & 0 & 0 \\
\textbf{RW SVM-C}  & \hlc7.6 & \hlc26 & \hlc68 & \hlc65 & 16 & 50 & 32 & 31 & 15 & 49 & 35 & 33 \\
\textbf{WW MLP-C}  & 13 & 42 & 57 & 57 & 24 & 73 & 23 & 26 & 16 & 52 & 47 & 48 \\
\textbf{WW SVM-C}  & 11 & 36 & 61 & 59 & 21 & 65 & 26 & 25 & 15 & 46 & 49 & 47 \\

\hline
\end{tabular}
\begin{tablenotes}\footnotesize
\item[] $^1$ Area (cm$^2$). $^2$ Power (mW). $^3$ Area and Power Gain compared to the bespoke baseline~\cite{Mubarik:MICRO:2020:printedml} (in $\%$).\vspace{-4pt}
\end{tablenotes}
\end{threeparttable}
\vspace{-2ex}
\end{table}

In Table~\ref{tab:threshold}, we consider a conservative accuracy loss threshold (i.e., only $1$\%) and we report the area and power values of the area-optimal circuits w.r.t. each approximation technique.
As shown, compared to the baseline bespoke~\cite{Mubarik:MICRO:2020:printedml} implementations, our cross-layer approximation delivers on average 47\% and 44\% area and power reduction, respectively.
The corresponding values of our coefficient approximation are 28\% and 26\%.
Standalone pruning achieves only 22\% and 20\% average area and power reduction, respectively.
Finally, to put these gains into perspective, we highlight in Table~\ref{tab:threshold} the circuits that could be power by one printed Molex $30$mW battery.
Overall, in addition to the circuits that even their exact bespoke baseline~\cite{Mubarik:MICRO:2020:printedml} can be power by a printed battery, our cross-layer approximation enables, for the first time, also printed MLP-C with $2$ hidden nodes and $34$ coefficients in total, printed MLP-R with $4$ hidden nodes and $48$ coefficients, and SVM-C with $15$ classifiers and $66$ coefficients.
On the other hand, compared with~\cite{Mubarik:MICRO:2020:printedml}, single layer approximation (only pruning or coefficient approximation) can not enable any additional circuits to be powered by a printed battery.

Finally, in Table~\ref{tab:exectime} we report the overall execution time of our framework.
As aforementioned due to the on-demand and point-of-use fabrication process, it is critical that the design time overhead due to approximation is kept to minimum.
As shown in Table~\ref{tab:exectime}, the average execution time of our framework is only $12$min while its median value is $8$min.
At the worst case, our framework required only $48$min for the Pendigits MLP-C, that is however very large to be considered for a printed circuit.
Note that the times in Table~\ref{tab:exectime} refer to the full design exploration (i.e., all the green dots in each subfigure of Fig.~\ref{fig:dse}).
As a result, the ``true'' Pareto front is obtained and the user may select a Pareto-optimal point based on the requirements (e.g., area-accuracy) of the printed application.

\section{Conclusion}\label{sec:conclusion}
Printed electronics prevail as a promising solution for a large number of application domains that require ultra low-cost, conformality, low time-to-market, nontoxicity, etc.
Nevertheless, large feature sizes in printed electronics prohibit the implementation of complex circuits.
In this work, we propose, for the first time, a novel cross-layer approximation framework that exploits peculiar printed electronics features and generates optimal approximate printed circuits for ML applications.
Our work demonstrates that approximate computing can enable the realization of several complex printed ML circuits.
\begin{table}[t!]
    \centering
    \footnotesize
    \caption{Execution Time of Our Framework in Minutes}
    \label{tab:exectime}
    \begin{threeparttable}
    \begin{tabular}{l|c|c|c|c}
    \hline
      & MLP-C & MLP-R & SVM-C & SVM-R \\ \hline
Cardio & 26 & 7 & 1 & 9 \\ \hline
Pendigits & 48 & -\tnote{1} & 14 & -\tnote{1} \\ \hline
RedWine & 7 & 6 & 2 & 7 \\ \hline
WhiteWine & 23 & 8 & 2 & 8 \\ \hline
    \end{tabular}
\begin{tablenotes}\footnotesize
\item[1] Not evaluated (see Table~\ref{tab:baselines})
\end{tablenotes}
\end{threeparttable}
\vspace{-4ex}
\end{table}

{\footnotesize
\section*{Acknowledgment}
This work is partially supported by the German Research Foundation (DFG) through the project ``ACCROSS: Approximate Computing aCROss the System Stack'' HE 2343/16-1.
}

{
\linespread{0.94}\selectfont
\bibliographystyle{IEEEtran}
\bibliography{references}
}

\end{document}